\documentclass{OAGM}
\OAGMarXiv{1505.01065}

\usepackage{amsmath}
\usepackage{abbrevs}
\usepackage{graphicx}
\usepackage{authblk}

\newname\eg{e.g.}
\newname\Eg{E.g.}
\newname\etc{etc.}
\newname\ie{i.e.}
\newname\etal{\em et al.}
\newname\cf{cf.}
\newname\Fig{Figure} % Fig.
\newname\Eq{Eq.}
\newname\Sec{Section} % Sec.
\newname\Tbl{Table} % Tab.

\def\reffig#1{\Fig \ref{#1}}
\def\reftbl#1{\Tbl \ref{#1}}
\def\refsec#1{\Sec \ref{#1}}
\def\refeq#1{\Eq(\ref{#1})}

\newcommand{\qm}[1]{``#1''}

\def\sizefigOneSmall{0.95} % 0.75
\def\vspacefigPre{0mm} % before a figure
\def\vspacefig{0mm} % between figure and caption

\def\vspacefigPost{0mm}

\definecolor{markcolor}{rgb}{1,0,0}

\graphicspath{{figures/}}

\title{Measuring Visibility using Atmospheric Transmission and Digital Surface Model}

\author[1]{Jean-Philippe Andreu}
\author[2]{Stefan Mayer}
\author[1]{Karlheinz Gutjahr}
\author[1]{Harald Ganster}

\affil[1]{ JOANNEUM RESEARCH Forschungsgesellschaft mbH\\
  DIGITAL - Institute for Information and Communication Technologies\\
  Steyrergasse 17, 8010 Graz, Austria}
\affil[2]{AUSTRO CONTROL \"Osterreichische Gesellschaft f\"ur Zivilluftfahrt mbH\\
  Schnirchgasse 11, 1030 Wien, Austria}

\begin{document}
\maketitle
% ---------------------------------------------------------------------------------------------------------------------
\begin{abstract}%\vspace{-2mm}
Reliable and exact assessment of visibility is essential for safe air traffic. In order to overcome the drawbacks of the currently subjective reports from human observers, we present an approach to automatically derive visibility measures by means of image processing. It first exploits image based estimation of the atmospheric transmission describing the portion of the light that is not scattered by atmospheric phenomena (e.g., haze, fog, smoke) and reaches the camera. Once the atmospheric transmission is estimated, a 3D representation of the vicinity (digital surface model: DMS) is used to compute depth measurements for the haze-free pixels and then derive a global visibility estimation for the airport. Results on foggy images demonstrate the validity of the proposed method.
\end{abstract}

% ---------------------------------------------------------------------------------------------------------------------
\section{Motivation}
In order to guarantee for safe air traffic controllers have to rely on precise forecasts and measurements of the current weather situation. Those have to be compiled and reported according to official regulations (e.g. visual flight rules (VFR) flight\footnote{Visual meteorological conditions: \url{http://en.wikipedia.org/wiki/Visual_meteorological_conditions}}) every 30 minutes in an international standard format \footnote{{METAR: M\'ET\'eorologique Aviation R\'eguli\`ere}}.
The decisions the controllers take do not only impact the security but also influence the air traffic and its economic repercussions in case of delays \cite{Allan2001}.

\begin{figure}[!h]
	\vspace{\vspacefigPre}
	\centering
	\resizebox*{\sizefigOneSmall\columnwidth}{!}{\includegraphics{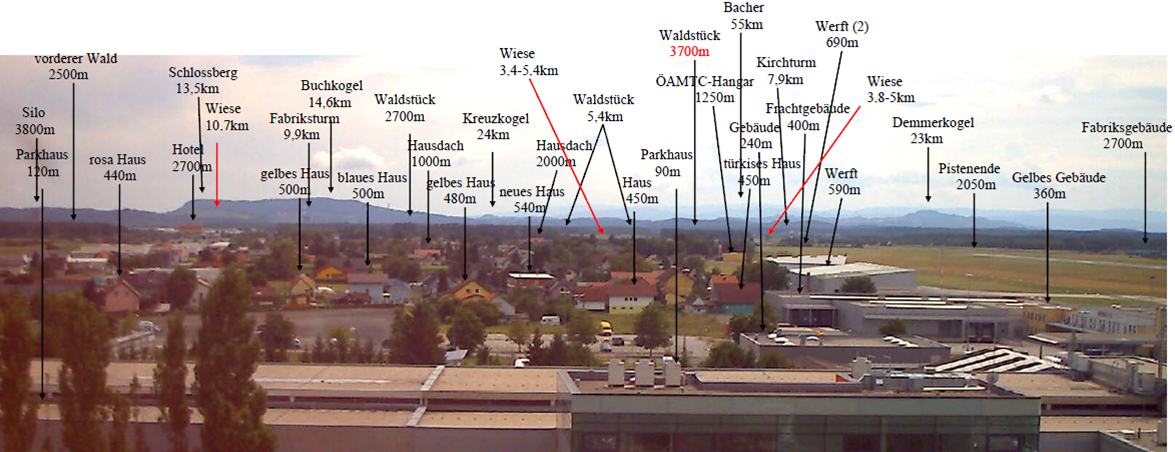}}
	\vspace{\vspacefig}
	\caption[]{Landmarks around the airport of Graz (Austria).}
  \label{fig:LandMarks}
	\vspace{\vspacefigPost}
\end{figure}

All major airports operate dedicated sensor systems to assess the current weather situation. Besides \qm{classic} parameters like wind, pressure, humidity, and temperature, there are point-like measurements of visibility and cloud cover information.
For reporting the prevailing visibility in the airport vicinity, observers currently compile their reports based on integrating sensor measurements with visual observation of known landmarks, like buildings, mountain tops, etc. (\reffig{fig:LandMarks}).
Human estimation is naturally very subjective to the individual observer and error-prone. 
Visibility sensors employ the principles of forward scattering to provide accurate and reliable information regarding visibility changes in the atmosphere. These sensors (usually located at the start and end of the runways) are very precise but only give very local (\eg at the position of the sensor) measurements.

Contrarily to Andreu et al. \cite{Andreu2011}, the approach presented in this paper is not aiming at emulating the controller procedure which actually results in a limited number of visibility measurements (up to the number of defined landmarks scattered around the airport) but instead intends to get a dense (\eg for each pixel in the image) visibility estimation. In order to quantify visibility, we propose a three steps procedure. First, we use a single image haze removal technique in order to recover the atmospheric transmission at every pixel of the image. Second, with the help of a very precise (\eg 1m resolution) digital surface model of the vicinity of the airport, a depth map is generated for the whole image. Finally, statistics about the depth of the scene points whose corresponding pixels are deemed visible gives the visibility in the field of view of the camera.

A short overview of the relevant literature in this field is presented in \refsec{sec:RelatedWork}, which is followed by the methods applied for estimating the atmospheric transmission in \refsec{sec:Transmission}. Then, it is shown in \refsec{sec:DepthMap} how, with the help of a very precise digital surface model of the airport vicinity, a depth map can be generated. In \refsec{sec:Visibility} we combine the estimated atmospheric transmission and the depth map to derive a global visibility estimation for the airport. Finally, \refsec{sec:Conclusion} concludes with final remarks.

% ---------------------------------------------------------------------------------------------------------------------
\section{Related work}
\label{sec:RelatedWork}
Weather and other atmospheric phenomena, such as haze, fog, mist, rain and snow, greatly reduce the visibility of distant regions in images of outdoor scenes. Quantifying the amount of atmospheric scattering or removing it (\eg dehazing or de-weathering), is a challenging problem, because the degree to which it effects each pixel depends on the depth of its corresponding scene point. This is leading to an under-constrained problem if the input is only a single image. Therefore, many methods propose using multiple images or additional information in order to alleviate that inherent problem.

Recent dehazing methods like \cite{Fattal2008} and \cite{Tan2008} are able to dehaze single images by making assumptions about the contrast difference in haze-free and hazy scenes as well as assumptions about the transmission and the surface shading. Even though those methods result in visually appealing images they are either not physically valid or fail in handling images with heavy haze.

In order to remove haze, Schechner \etal \cite{Schechner2001} successfully made use of multiple images taken with different polarizer orientations, which is difficult to adapt for standard webcam single images.

Depth-based methods, where depth information or heuristics about the scene are provided either by interactive user inputs or by photo realistic 3D models, have been successful in de-weathering images. Narasimhan \etal \cite{Narasimhan2003} addressed the question of de-weathering a single image using simple additional information (\ie sky, vanishing point) provided interactively by the user. Kopf \etal \cite{Kopf2008} take advantage of the availability of accurate 3D models and try to estimate stable values for the haze function directly from the relationship between the colors in the image and those of the rendered 3D model. However, in our case, even if we do have access to a pretty precise digital surface model generated by laser scanning, this 3D model has no color information thus preventing the use of Kopf \etal's approach.

% ---------------------------------------------------------------------------------------------------------------------
\section{Atmospheric Transmission}
\label{sec:Transmission}
\subsection{Dark Channel Prior}
In computer vision, the model widely used to describe the formation of an image perturbed by haze is \cite{Narasimhan2002}, \cite{Narasimhan2000}, \cite{Tan2008}:
\begin{equation} \label{eq:HazeImage}
I\left(x\right)=J\left(x\right)t(x)+A\left(1-t(x)\right)
\end{equation}
where $I$ is the observed intensity, $J$ is the scene radiance, $A$ is the global atmospheric light, and $t$ is the atmospheric transmission describing the portion of the light that is not scattered and reaches the camera. The first term $J(\boldsymbol{x})t(\boldsymbol{x})$ on the right-hand side is called direct attenuation, and the second term $A(1-t(\boldsymbol{x}))$ is called air-light. When the atmosphere is homogenous, the transmission $t$ is attenuated exponentially with the scene depth and can can be expressed as $t(x) = e^{-\beta d(x)}$ where $\beta$ is the scattering coefficient of the atmosphere and $d$ the scene depth. 

\begin{figure}[!h]
	\centering
	\null\hfill
		\includegraphics[width=0.3\hsize]{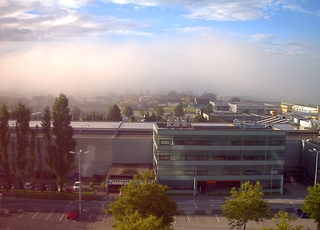}
	\hfill
		\includegraphics[width=0.3\hsize]{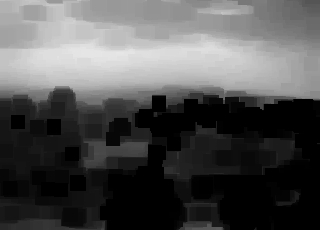}
	\hfill
		\includegraphics[width=0.3\hsize]{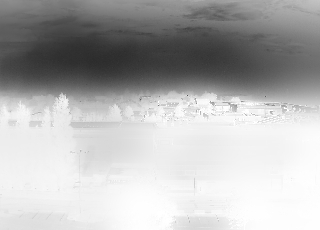}
	\hfill\null
	\caption{Foggy image, its dark channel computed over 15x15 patches, and transmission map after guided filtering with the source image.}
	\label{fig:DarkChannel}
\end{figure}

He \etal \cite{He2011} propose using a \textit{dark channel prior} for single image haze removal. This prior is partially inspired by the well-known dark-object subtraction technique \cite{Chavez1988} used in multispectral remote sensing systems.
The idea is that in most of the local regions which do not cover the sky, some pixels (called dark pixels) normally have very low intensity in at least one color (RGB) channel. In hazy images, the intensity of these dark pixels in that channel is mainly contributed by the air-light. Therefore, these dark pixels can directly provide an accurate estimation of the haze transmission.
For an arbitrary image $I$, its dark channel $I^{dark}$ is given by $I^{dark}\left(x\right)=\underset{y\in \Omega (x)}{\text{min}}\left(\underset{c\in \left\{r,g,b\right\}}{\text{min}}I^c(y)\right)$ where $I^c$ is a colour channel of $I$ and  is a local patch centred at $x$.
Using the concept of a dark channel, if $I$ is an outdoor haze-free image, except for the sky region, the intensity of $I$'s dark channel is low and tends to be zero. \reffig{fig:DarkChannel} shows a foggy image of the airport surrounding (left) and its corresponding dark channel computed over local patches of size 15x15 pixels (center).

\subsection{Estimating the transmission}
As in \cite{He2011}, $A$ is automatically estimated from the most haze-opaque pixels (the top 0.1\% brightest pixels in the dark channel). If the transmission $t(x)$ can be assumed constant (denoted as $\tilde{t}(x)$ ) over the local patch $\Omega(x)$,  \refeq{eq:HazeImage}, normalized by $A$ for each colour channel, could be rewritten as:
\begin{equation} \label{eq:NormalizedHazeImage}
\underset{y\in \Omega (x)}{\text{min}}\left(\underset{c\in \left\{r,g,b\right\}}{\text{min}}\frac{I^c(y)}{A^c}\right)=\tilde t(x)\underset{y\in \Omega (x)}{\text{min}}\left(\underset{c\in \left\{r,g,b\right\}}{\text{min}}\frac{J^c(y)}{A^c}\right)+1-\tilde t(x)
\end{equation}
By definition the scene radiance $J$ should be haze-free, its dark channel is close to zero so \refeq{eq:NormalizedHazeImage} leads to:
\begin{equation} \label{eq:JtoZero}
\tilde t(x)=1-\underset{y\in \Omega (x)}{\text{min}}\left(\underset{c\in \left\{r,g,b\right\}}{\text{min}}\frac{I^c(y)}{A^c}+1\right)
\end{equation}
where $\underset{y\in \Omega (x)}{\text{min}}\left(\underset{c\in \left\{r,g,b\right\}}{\text{min}}\frac{I^c(y)}{A^c}+1\right)$ is the dark channel of the normalized hazy image. In practice, even on clear days the atmosphere is not absolutely free of any particle meaning that haze still exists when we look at distant objects. So, a very small amount of haze could be optionally kept for distant objects by introducing a constant parameter $\omega (0< \omega \leq 1) $ into \refeq{eq:JtoZero}:
\begin{equation} \label{eq:Transmission}
\tilde  t(x)=1-\omega \underset{y\in \Omega (x)}{\text{min}}\left(\underset{c\in \left\{r,g,b\right\}}{\text{min}}\frac{I^c(y)}{A^c}+1\right)
\end{equation}
Note that this method for estimating the atmospheric transmission is based on a prior (Dark Channel) reflecting the statistics of outdoor images. That prior may become invalid on objects similar to the atmospheric light (\ie white walls) or having no shadow cast on them. As a result, this method might underestimate the transmission of these objects and overestimate the haze layer. However, images of the vicinity of an airport do not contain many of such problematic objects and have plenty enough shadow casting.

\subsection{Refining the transmission}
Using \refeq{eq:Transmission} for computing the transmission results in a very coarse map. Since the transmission is not always constant over local $\Omega (x)$ patches, the map has to be refined in order to capture the sharp edge discontinuities and outline the profile of the objects.
We follow the approach of He \etal \cite{He2011} applying a guided filter (with the original hazy image used as the guidance image) to refine the transmission map \cite{He2013}. \reffig{fig:DarkChannel} (right) shows the corresponding refined version of the transmission map.

% ---------------------------------------------------------------------------------------------------------------------
\section{Depth map generation}
\label{sec:DepthMap}
\begin{figure}[!h]
  \centering
	\includegraphics[height=3.6cm]{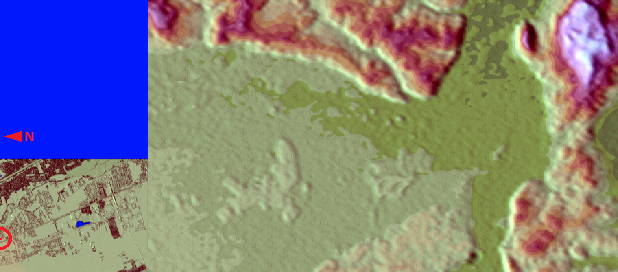}
	\caption{Painted relief of the combined LiDAR SRTM DSM south to the tower (position outlined at the center of the red circle in the lower left part of the image) of the airport of Graz (Austria). Blue areas indicate no data. }
  \label{fig:LIDARRelief}
\end{figure}

\begin{figure}[!h]
  \centering
	\includegraphics[height=3.6cm]{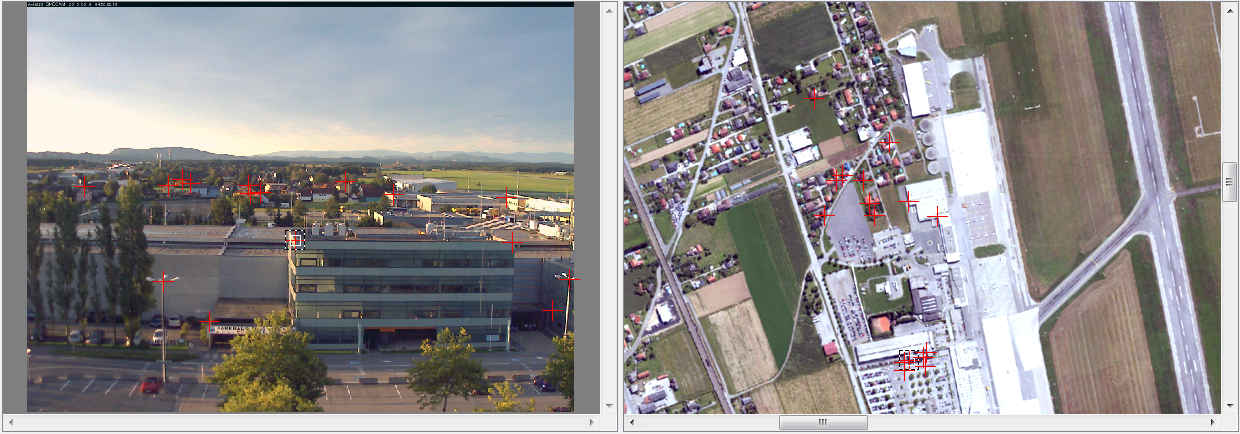}
	\caption{Identification of GCPs in one image of the webcam of the airport Graz (Austria) (left) and combined LiDAR-SRTM and Google map (right).}
  \label{fig:GoogleMap}
\end{figure}

For this study we identified 16 ground control points (GCPs) manually for a part (\ie~camera field of view) of the airport vicinity (\reffig{fig:LIDARRelief}). The 3D reference coordinates of the GCPs were derived from a combined LiDAR\footnote{Light Detection and Ranging}-SRTM\footnote{Shuttle Radar Topography Mission} DSM. In the northern part in the vicinity of the tower of the airport Graz we therefore used a LiDAR DSM which was acquired with 4pts per sqm in the year 2009. For more distant areas we used the globally available version 4 of the SRTM DSM, which was acquired using single pass interferometry at C-band with a ground resolution of about 90m in the year 2000 \cite{CGIAR}. \reffig{fig:LIDARRelief} shows a painted relief of the generated combined LiDAR SRTM DSM. To facilitate the identification of the buildings given in the LiDAR DSM we additionally used Google map (See \reffig{fig:GoogleMap}) and here with especially oblique views.

\begin{figure}[!h]
  \centering
	\null\hfill
		\includegraphics[height=3.6cm]{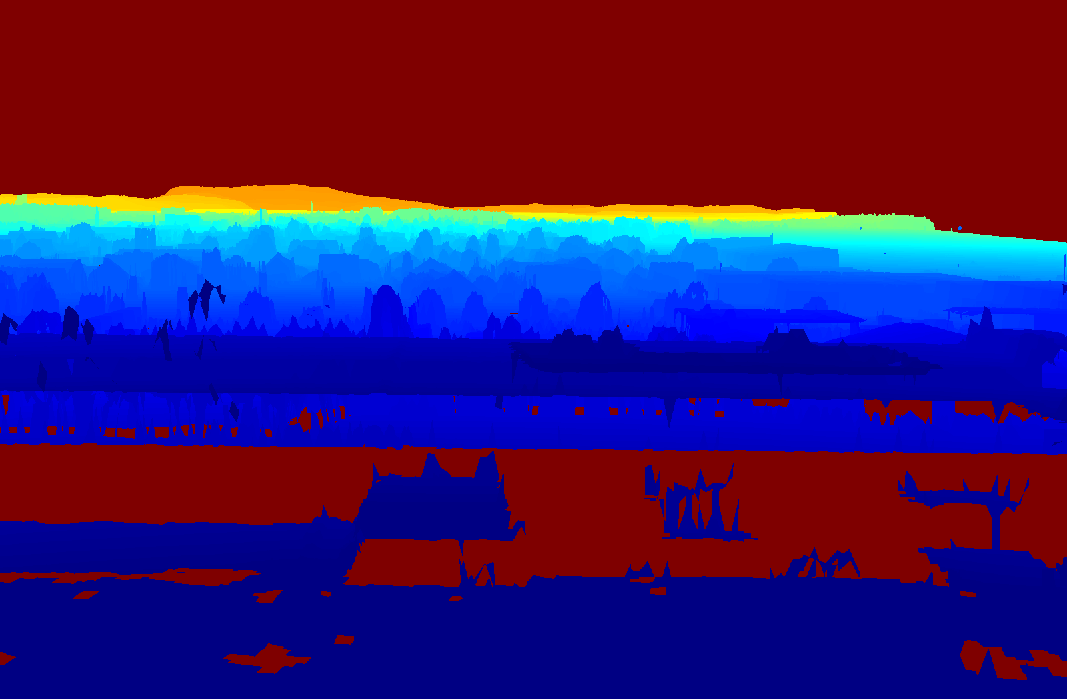}
	\hfill
		\includegraphics[height=3.6cm]{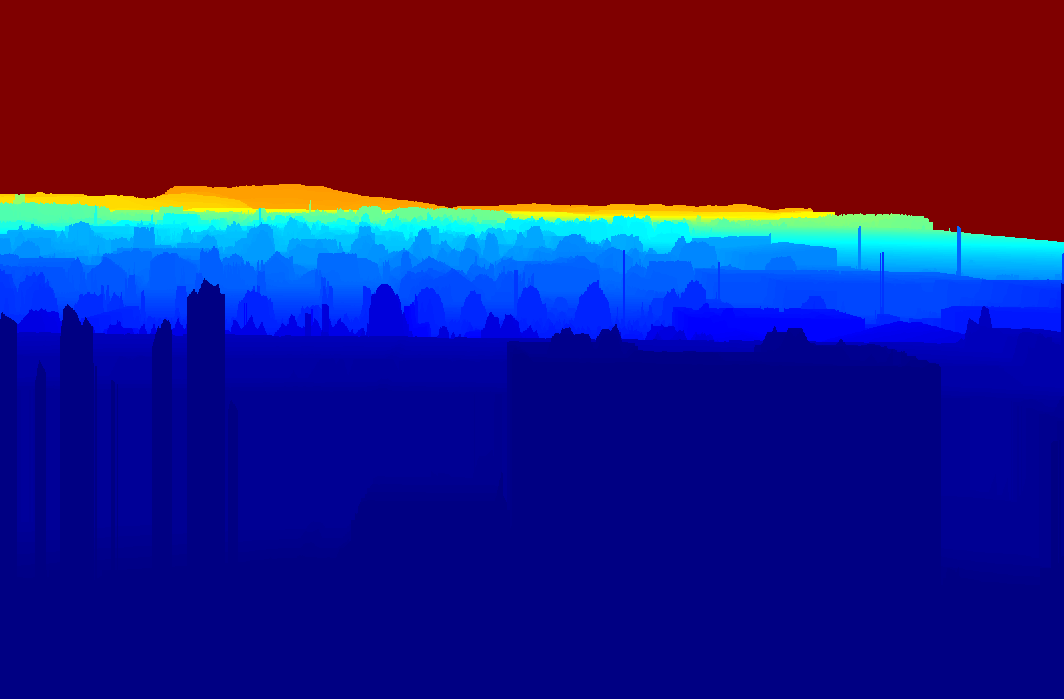}
	\hfill\null
	\caption{Raw depth map of the airport Graz (Austria) (left) and corrected depth map (right). Depths are given in logarithmic scale from 60m up to 16km.}
	\label{fig:DepthMap}
\end{figure}

Based on the well-known (linearized) co-linearity equations we used the GCP information to orient this image \cite{Kraus2007}. Besides the extrinsic orientation parameters (camera position and rotation) we limited the intrinsic parameters to the focal length only. For a more precise distance map also the principal point and camera distortion parameters could be derived. A coarse position of the webcam and the knowledge that the camera is pointing to the South were sufficient to set-up the orientation with a root mean square error of less than 4 pixels (with respect to the GCPs reprojection error).
Once the image is oriented, for every pixel of the image the first intersection of the line-of-sight (LOS) with the combined LiDAR-SRTM DSM is calculated and the distance assigned to the raw distance map. More precise, we restrict ourselves only to image pixels below the detected horizon. As the combined LiDAR-SRTM DSM is only 2.5 dimensional this calculation fails if the LOS is below e.g. the roof of a building. In this case the intersection is found with an object behind the actual building. We therefore applied a simple reconstruction of facades by projecting the eaves line to the ground. Then a simple search in the image line direction ensures that the range direction is always decreasing (when starting from the horizon). \reffig{fig:DepthMap} shows the raw distance map of the airport Graz (Austria) and corrected distance map. All distances are given in logarithmic scale ranging from less than 60 m to more than 16 km.

% ---------------------------------------------------------------------------------------------------------------------
\section{Visibility estimation}
\label{sec:Visibility}
\begin{figure}[!h]
  \centering
		\includegraphics[height=4.2cm]{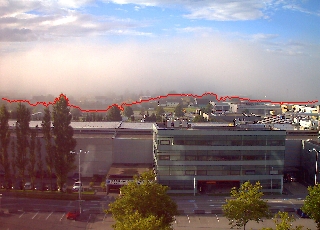}
	\hfill
		\includegraphics[height=4.2cm]{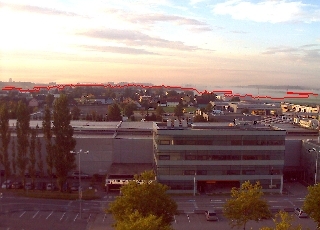}
	\caption{Highlighted (in red) haze border by using a global threshold of 0.75 on the corresponding transmission maps.}
	\label{fig:VisibilityDelineation}
\end{figure}

\reffig{fig:VisibilityDelineation} shows the result of detecting haze on two hazy images taken at different day time under different weather conditions (low-level clouds and fog banks). The border of the haze is delineated by finding the boundaries of the region obtained by thresholding the transmission map computed using the methods described in \refsec{sec:Transmission}.

As can be seen on the both examples of \reffig{fig:VisibilityDelineation}, the border between hazy and non hazy regions of the image is delineated by using a threshold of 0.75 (heuristically estimated) on the corresponding transmission maps. On the right hand side, one can notice that the fog bank over the runway is accurately segmented which is of importance for air traffic.
Derivating a visibility distance estimation out of the depth and transmission map is done by statistic analysis on the pixels of the depth map whose transmission value in the corresponding transmission map is above a given threshold value. One can then either choose as visibility distance the overall depth maximum of the concerned pixels or, in order to eventually avoid outliers, a simple percentile statistics over the gathered depths (\ie the top 1\%).

\reftbl{table:TestVisibility} reports visibility distance estimations from an initial evaluation together with the Austro Control observers. It is done with a global threshold value of 0.75 on the transmission map and the distance corresponding to the 99th percentile rank of the depth map histogram (with a bin distance of 10m) as resulting visibility distance.
\begin{table}[!h]
\centering
\scriptsize{
\begin{tabular}{||c|c||c|c||c|c||}
\hline  Measured & Reported & Measured & Reported & Measured & Reported \\
                 & (METAR)  &          & (METAR)  &          & (METAR)  \\
\hline
\hline 102m & 250m & 731m & 500m & 6768m & 6000m \\
\hline 110m & 300m & 443m & 600m & 10537m & 10km \\
\hline 371m & 300m & 791m & 800m & 12178m & 10km \\
\hline
\end{tabular}}
\caption{Comparison of measured to officially reported visibility (METAR).}
\label{table:TestVisibility}
\end{table}
Globally, the results over a broad range of distances (from as close as 250m up to 10km) are mimicking official reports well. Some discrepancies could be explained by the difficulty for the controller to evaluate the visibility distance due to the low number of landmarks in some distance ranges (\eg in the range [600m, 800m], cf. \reffig{fig:LandMarks}) and also due to the subjectivity of the concept of visibility as human beings are defining it (\eg when is landmark visible and when not?). Clearly, the value heuristically chosen as global threshold on the transmission has a major impact on the visibility results. No criticality analysis has been done on this parameter yet since the results presented here are mostly preliminary and mainly thought as a proof of concept.

% ---------------------------------------------------------------------------------------------------------------------
\section{Conclusion}
\label{sec:Conclusion}
In this paper, we have proposed an approach to automatically derive visibility measures at airport sites by combining an image processing method with the digital surface model of the airport vicinity. The image processing method equivalent to an image haze removal technique enables us to first recover the atmospheric transmission at every pixel in the image. Then, with the help of a very precise digital surface model of the airport vicinity, a depth map is generated for every pixel. Finally, gathering the depth of the pixels which are visible gives the visibility in the field of view of the camera. It was shown that this approach gives preliminary results comparable to the visibility estimations given by human controllers. Further work has to be done in order to automatically get an optimal value of the transmission threshold (\ie by exploiting local features for threshold derivation) which is a critical parameter of the proposed approach. Also, instead of being global, we will investigate threshold values that could be locally adapted to different parts of the image.

\subsubsection*{Acknowledgments.}
This research was supported by the TAKE OFF program, an initiative of the Austrian \qm{Federal Ministry for Transport, Innovation and Technology} under contract number 843985.

\bibliography{2015_oagm}

\end{document}